\documentclass[conference]{IEEEtran}
\IEEEoverridecommandlockouts
\usepackage{cite}
\usepackage{amsmath,amssymb,amsfonts}
\usepackage{algorithm}
\usepackage[noend]{algpseudocode}
\usepackage{graphicx}
\usepackage{textcomp}
\usepackage{xcolor}

\usepackage[colorlinks=true, citecolor=blue, linkcolor=.]{hyperref}
\usepackage{cleveref}

\usepackage{placeins}
\usepackage{pdfpages}
\usepackage{booktabs}
\usepackage[most]{tcolorbox}
\usepackage{calligra}
\usepackage{caption}
\usepackage{datetime}
\usepackage{dblfnote}
\usepackage{dirtytalk}
\usepackage{dsfont}
\usepackage{fancyhdr}
\usepackage{fix-cm}
\usepackage[T1]{fontenc}
\usepackage{textcomp,gensymb} 
\usepackage{graphicx}
\usepackage{lipsum}
\usepackage{listings}
\usepackage{transparent}
\usepackage[everyline=true,framemethod=tikz]{mdframed}
\usepackage{mparhack}
\usepackage{multicol}
\usepackage{multirow}
\usepackage{parskip}
\usepackage{lscape}
\usepackage{pdflscape}
\usepackage{pdfpages}
\usepackage{placeins}
\usepackage{rotating}
\usepackage{setspace}
\usepackage{subcaption}
\usepackage{threeparttable}
\usepackage[normalem]{ulem}
\usepackage{verbatim}
\usepackage{soul} 

\usepackage{array}
\newcolumntype{L}[1]{>{\raggedright\let\newline\\\arraybackslash\hspace{0pt}}m{#1}}
\newcolumntype{C}[1]{>{\centering\let\newline\\\arraybackslash\hspace{0pt}}m{#1}}
\newcolumntype{R}[1]{>{\raggedleft\let\newline\\\arraybackslash\hspace{0pt}}m{#1}}

\algrenewcommand\algorithmicforall{\textbf{foreach}}
\algrenewcommand\algorithmicindent{.8em}

\usepackage{blindtext}
\usepackage{lastpage}
\usepackage{fancyhdr}
\usepackage{censor}

\def\BibTeX{{\rm B\kern-.05em{\sc i\kern-.025em b}\kern-.08em
    T\kern-.1667em\lower.7ex\hbox{E}\kern-.125emX}}
\begin{document}

\title{Embodied AI in Mobile Robots: Coverage Path Planning with Large Language Models}

\StopCensoring

\author{
\IEEEauthorblockN{\censor{Xiangrui Kong}}
\IEEEauthorblockA{
\censor{\textit{Dept. of Electrical, Electronic 
and Computer Engineering}} \\
\censor{\textit{The University of Western Australia}} \\
\censor{Perth, Australia} \\
\censor{xiangrui.kong@research.uwa.edu.au}}

\and

\IEEEauthorblockN{\censor{Wenxiao Zhang}}
\IEEEauthorblockA{
\censor{\textit{Dept. of Computer Science and Software Engineering}} \\
\censor{\textit{The University of Western Australia}} \\
\censor{Perth, Australia} \\
\censor{wenxiao.zhang@research.uwa.edu.au}}

\and

\IEEEauthorblockN{\censor{Jin Hong}}
\IEEEauthorblockA{
\censor{\textit{Dept. of Computer Science and Software Engineering}} \\
\censor{\textit{The University of Western Australia}} \\
\censor{Perth, Australia} \\
\censor{jin.hong@uwa.edu.au}}

\and

\IEEEauthorblockN{\censor{Thomas Braunl}}
\IEEEauthorblockA{
\censor{\textit{Dept. of Electrical, Electronic and Computer Engineering}} \\
\censor{\textit{The University of Western Australia}} \\
\censor{Perth, Australia} \\
\censor{thomas.braunl@uwa.edu.au}}
}

\maketitle

\begin{abstract}
In recent years, Large Language Models (LLMs) have demonstrated remarkable capabilities in understanding and solving mathematical problems, leading to advancements in various fields. We propose an LLM-embodied path planning framework for mobile agents, focusing on solving high-level coverage path planning issues and low-level control.
Our proposed multi-layer architecture uses prompted LLMs in the path planning phase and integrates them with the mobile agents' low-level actuators. To evaluate the performance of various LLMs, we propose a coverage-weighted path planning metric to assess the performance of the embodied models. Our experiments show that the proposed framework improves LLMs' spatial inference abilities.
We demonstrate that the proposed multi-layer framework significantly enhances the efficiency and accuracy of these tasks by leveraging the natural language understanding and generative capabilities of LLMs. Our experiments show that this framework can improve LLMs' 2D plane reasoning abilities and complete coverage path planning tasks.
We also tested three LLM kernels: \textit{gpt-4o}, \textit{gemini-1.5-flash}, and \textit{claude-3.5-sonnet}. The experimental results show that \textit{claude-3.5} can complete the coverage planning task in different scenarios, and its indicators are better than those of the other models.

\end{abstract}


\section{Introduction} 

\begin{figure*}[!h]
    \centering
    \includegraphics[width=\textwidth]{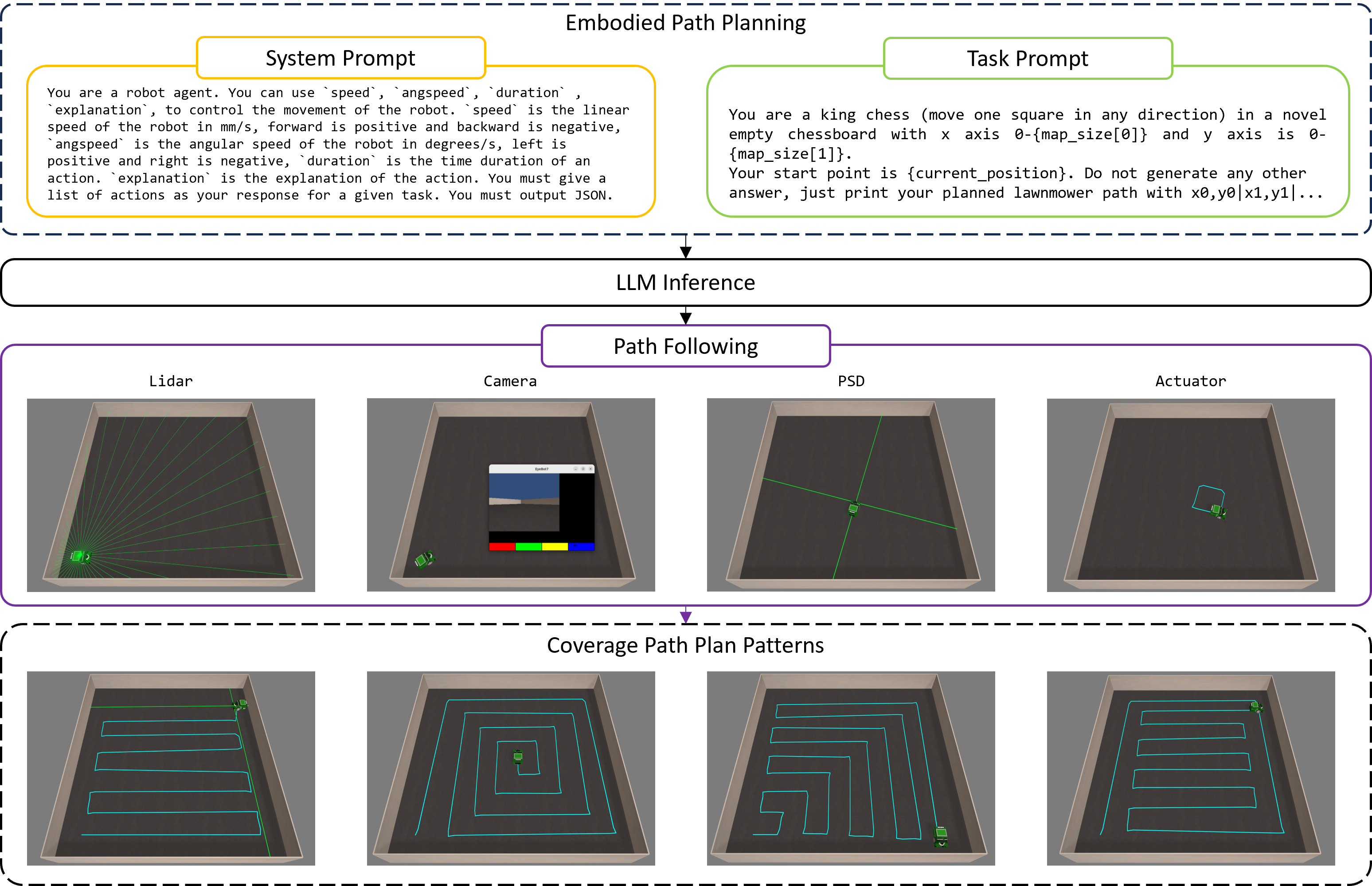}
    \caption{Multi-layer embodied path planning  framework}
    \label{fig:multilayer_framework}
\end{figure*}

The application of Large Language Models (LLMs) has grown exponentially, revolutionizing various fields with their advanced capabilities \cite{hadi2023survey}. Modern LLMs have evolved to perform various tasks beyond natural language processing. When integrated into mobile agents, these LLMs can interact with the environment and perform tasks without the need for explicitly coded policies or additional model training. This capability leverages the extensive pre-training of LLMs, enabling them to generalize across tasks and adapt to new situations based on their understanding of natural language instructions and contextual cues.

Embodied AI refers to artificial intelligence systems integrated into physical entities, such as mobile robots, that interact with the environment through sensors and actuators \cite{chrisley2003embodied}. The integration of LLMs with embodied AI in applications such as autonomous driving \cite{dorbala2024can} and humanoid robots \cite{cao2024ai} demonstrates their potential. However, the application of LLMs in controlling mobile robots remains challenging due to issues such as end-to-end control gaps, hallucinations, and path planning inefficiencies. LLMs possess the capability to solve mathematical problems, which directly aids in path planning methods \cite{gullms}.

Path planning and obstacle avoidance are critical for the effective operation of mobile robots, ensuring safe and efficient navigation in dynamic environments \cite{hewawasam2022past}. Coverage path planning is a typical method employed in various research areas, such as ocean seabed mapping \cite{galceran2012efficient}, terrain reconstruction \cite{TORRES2016441}, and lawn mowing \cite{hazem2021design}.
Traditional path planning methods include algorithms such as A* \cite{warren1993fast}, D* \cite{ferguson2005field}, and potential field methods \cite{barraquand1992numerical}. Given a global map, a path-planning method can be framed as a mathematical problem solvable by LLMs. In this context, we simplify some traditional path-planning methods and test LLMs in our mobile robot simulator. LLMs demonstrate their ability to solve mathematical problems collaboratively \cite{zhang2024mathverse}.

This paper presents a multi-layer coverage path planner based on existing multimodal large language models. It involves the static low-dimensional deconstruction of unstructured maps, abstracting spatial relationships into mathematical problems for reasoning and solving by prompted LLMs. The reasoning accuracy of the LLM is enhanced through multi-turn dialogues and multimodal interactions. The inferred results from the LLM are combined with the control interface, enabling the mobile agent to control the robot in real time for path planning. Simulation experiments demonstrate that LLMs possess path-planning capabilities in unstructured static maps.

\section{Related works} 

\subsection{LLMs in mobile robots}
Currently, LLMs are involved in various aspects of mobile robots, including code writing, model training, action interpretation, and task planning.
LLMs can process new commands and autonomously re-compose API calls to generate new policy code by chaining classic logic structures and referencing third-party libraries \cite{liang2023code}.
LLMs have also been used to automatically generate reward algorithms for training robots to learn tasks such as pen spinning \cite{ma2024eureka}.
PaLM-E, an embodied language model trained on multi-modal sentences combining visual, state estimation, and textual input encodings, demonstrates the versatility and positive transfer across diverse embodied reasoning tasks, observation modalities, and embodiments \cite{driess2023palme}.
LLMs have shown promise in processing and analyzing massive datasets, enabling them to uncover patterns, forecast future occurrences, and identify abnormal behaviour in a wide range of fields \cite{su2024large}.
VELMA is an embodied LLM agent that generates the next action based on a contextual prompt consisting of a verbalized trajectory and visual observations of the environment \cite{velma}.
Sharma et al. propose a method for using natural language sentences to transform cost functions, enabling users to correct goals, update robot motions, and recover from planning errors, demonstrating high success rates in simulated and real-world environments \cite{sharma2022correcting}.

There is also some research applying LLMs in zero-shot path planning.
The 3P-LLM framework highlights the superiority of the GPT-3.5-turbo chatbot in providing real-time, adaptive, and accurate path-planning algorithms compared to state-of-the-art methods like Rapidly Exploring Random Tree (RRT) and A* in various simulated scenarios \cite{latif20243pllm}.
Singh et al. describe a programmatic LLM prompt structure that enables the generation of plans functional across different situated environments, robot capabilities, and tasks \cite{singh2022progprompt}.
Luo et al. demonstrate the integration of a sampling-based planner, RRT, with a deep network structured according to the parse of a complex command, enabling robots to learn to follow natural language commands in a continuous configuration space \cite{kuo2020deep}.
ReAct utilizes LLMs to generate interleaved reasoning traces and task-specific actions \cite{yao2022react}.
These methods typically use LLMs to replace certain components of mobile robots. The development of a hot-swapping path-planning framework centred around LLMs is still in its early stages.

\subsection{Path planning method}
Path planning for mobile robots involves determining a path from a starting point to a destination on a known static map \cite{ab2024improved}. 
Obstacle avoidance acts as a protective mechanism for the robot, enabling interaction with obstacles encountered during movement. Low-level control connects algorithms to different types of system agents, such as UAVs, UGVs, or UUVs \cite{ali2024motion}. 
In addition to the A* and D* algorithms mentioned in the previous chapter, path planning algorithms include heuristic optimization methods based on pre-trained weights, such as genetic algorithms \cite{castillo2007multiple}, particle swarm optimization \cite{dewang2018robust}, and deep reinforcement learning \cite{panov2018grid}. 
These pre-trained methods do not directly rely on prior knowledge but utilize data to pre-train weights. 
The obstacle avoidance problem addresses dynamic obstacles encountered during movement, ensuring the safety of the mobile agent. 
Mainstream methods include the Artificial Potential Field.

\begin{figure}[!t]
	\centering
    \captionsetup[subfloat]{labelfont=scriptsize,textfont=scriptsize}
	\subfloat[Standard lawnmower]{\includegraphics[width=1.6in,height=1.1in]{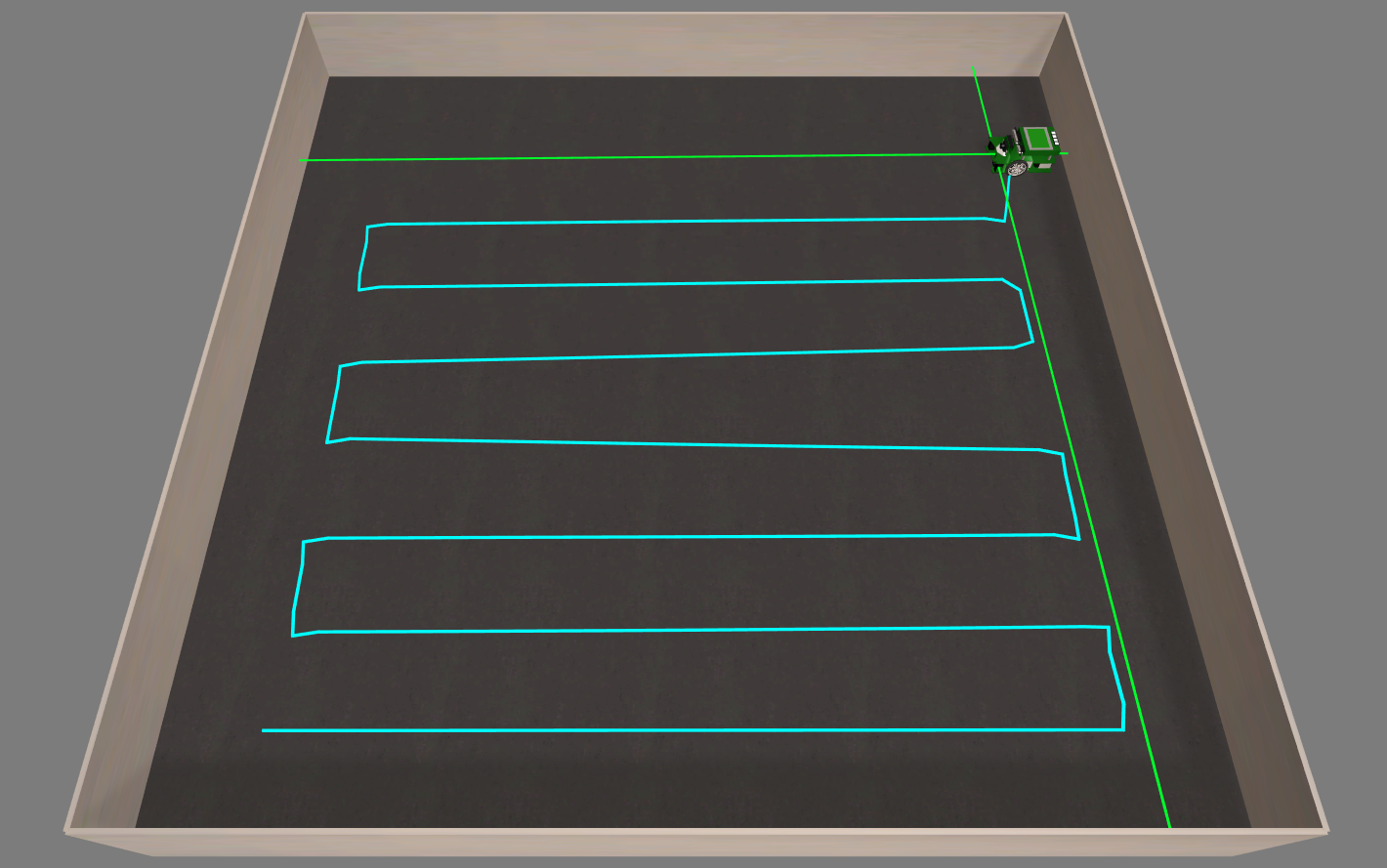}%
		\label{fig:pt1}}
	\hfil
	\subfloat[Square spiral]{\includegraphics[width=1.6in,height=1.1in]{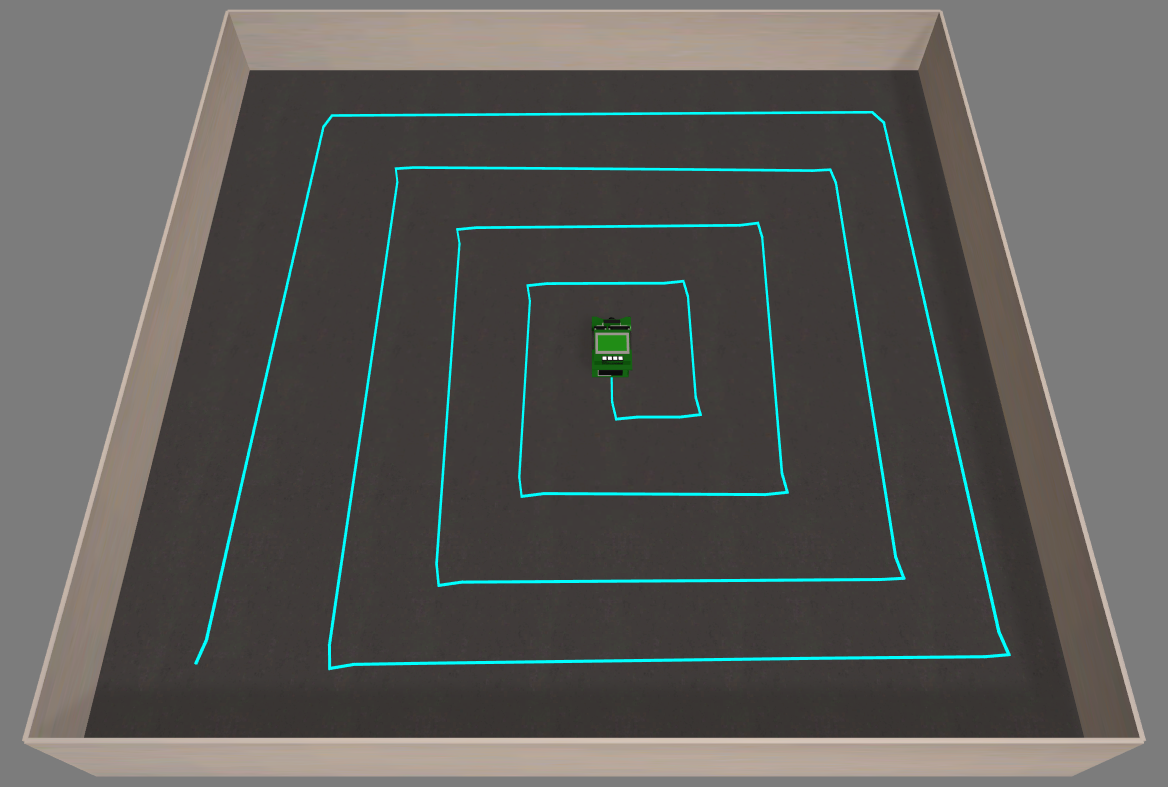}%
		\label{fig:pt2}}
	\hfil
	\subfloat[Square move]{\includegraphics[width=1.6in,height=1.1in]{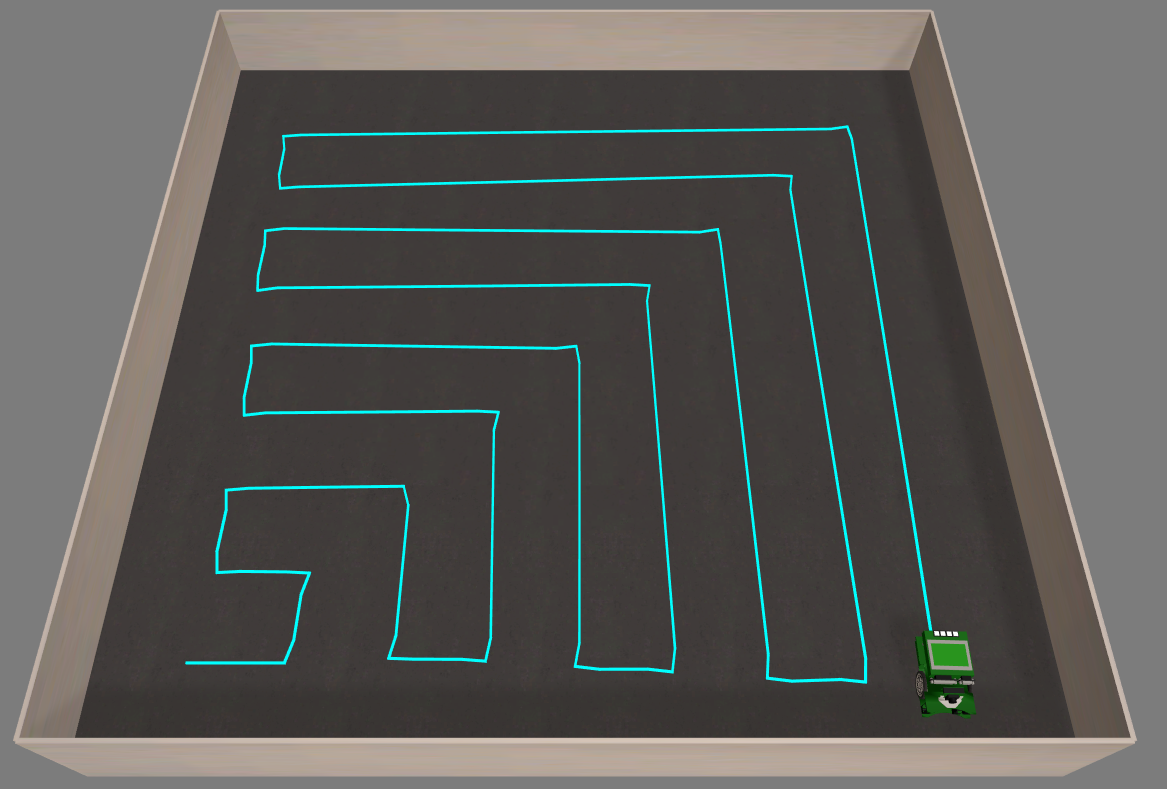}%
		\label{fig:pt3}}
	\hfil
	\subfloat[Lawnmower after wall following]{\includegraphics[width=1.6in,height=1.1in]{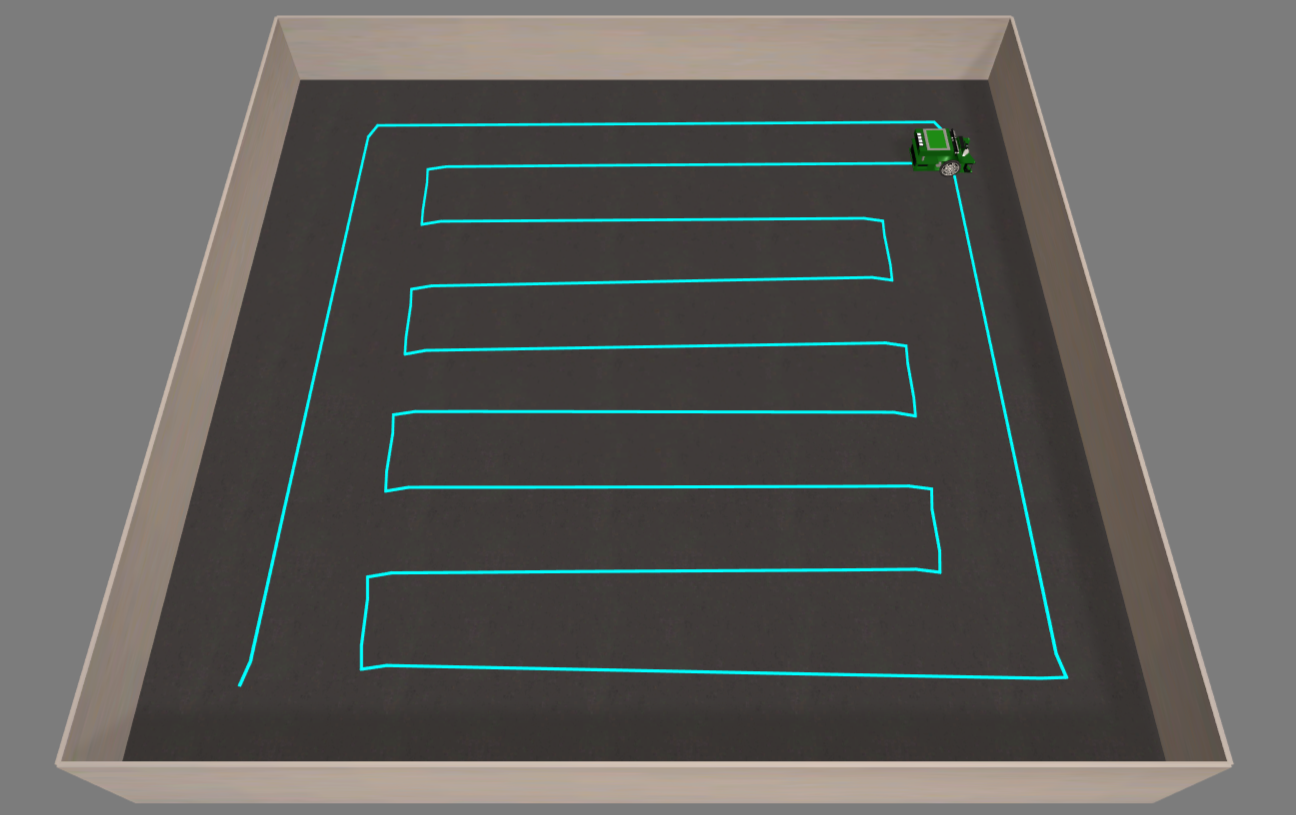}%
		\label{fig:pt4}}
    \caption{Comparison of Path Planning Patterns generated by prompted LLMs}
	\label{fig:pattern}
\end{figure}

The coverage path planning problem is a branch of path planning problems. Compared with point-to-point path planning, a coverage waypoint list needs to cover the given area as much as possible \cite{di2016coverage}. 
Classically, decomposing a given map based on topological rules and then applying a repeatable coverage pattern is a common way to solve this issue following the divide-and-conquer algorithm \cite{saguesmulti,petitjean2002survey,smith1985design}. 
In this way, a known map is required to start, whereas the Traveling Salesman Problem (TSP), an optimization problem that seeks to determine the shortest possible route for a salesman to visit a given set of cities exactly once and return to the original city, offers another solution to solve it in a node graph \cite{hoffman2013traveling}.
Figure~\ref{fig:pattern} presented showcases a comparison of four distinct path-planning patterns employed in robotic navigation. The first pattern, labelled as a standard lawnmower (Figure~\ref{fig:pt1}), utilizes a standard back-and-forth sweeping motion to ensure comprehensive coverage of the area. The second pattern, square spiral (Figure~\ref{fig:pt2}), depicts a robot following an inward spiral trajectory, efficiently covering the space in a continuous inward motion. The third pattern, square move (Figure~\ref{fig:pt3}), illustrates a robot navigating in a sequential inward square formation, progressively moving towards the centre. Finally, the lawnmower after wall following (Figure~\ref{fig:pt4}) combines two approaches: initially, the robot adheres to the perimeter of the wall following area, and subsequently, it adopts a lawnmower pattern to cover the remaining interior space. This comparative analysis of path planning strategies highlights the versatility and application-specific advantages of each method in ensuring thorough area coverage in robotic navigation tasks.

\section{Methodology}
As depicted in Figure~\ref{fig:multilayer_framework}, our method is divided into three main sections: global planning, waypoint evaluation, and navigation. 
In \textit{global planning} phase, a coverage planning task in a given map is decomposed into a cell map, and the additional requirement is designed using natural language with a simplified format to decompose LLM responses. 
During the \textit{waypoint evaluation} phase, the LLM responses are further evaluated before execution.
The theoretical coverage rate and the theoretical shortest path distance are calculated in this phase.
Once the desired path passes the evaluation, the planned warpaint list transitions to the navigation phase.
In \textit{navigation} phase, the mobile agent simply travels through them one by one and triggers the safety mechanism if the sensor shows a threshold distance between the robot and an unknown obstacle.

\subsection{Global planning}
We design a waypoint generation prompt with natural language describing 2D grid maps like a chessboard to simplify the inference difficulty of LLMs. 
During the global phase, a prompt contains the size of the grid map, current location, and response format. 
We assume the LLM generates the desired waypoint list with a required format which is a local position sequence separated with a bar sign.
In order to evaluate the performance and excitability of the planned path, the desired waypoint list is visualised and calculated in the phrase of waypoint evaluation.
Considering the robot's kinematic limitation, we prompt a description of mobile agents including equipped sensors, driving commands, and basic status. 
We experimented with various settings to describe robot behaviors in conversations with ChatGPT. However, we observed that these changes in description had minimal impact on the output responses. 
We use OpenAI GPT-4o services \cite{achiam2023gpt}, a multimodal efficient model for inference and reasoning.
The temperature parameter with the range from $0$ to $2$ is set as $0.6$ with our prompt for a consistent planned path.
Lower values for temperature result in more consistent outputs, while higher values generate more diverse and creative results.

\subsection{Waypoint evaluation}
The response from the LLMs can occasionally be incorrect, leading us to design a waypoint evaluator to mitigate hallucinations. 
Initially, the desired waypoint list is visualized on a 2D map, providing a clear and precise layout of the proposed route. 
The shortest path and the number of turns are then calculated mathematically to ensure efficiency and feasibility. 
Paths that do not meet the required criteria are rejected and not converted into a driving command list. 
The designed dialogue system initiates as soon as the agent receives the task command and map, continuing until a waypoint list passes the evaluation. 
This ensures that only optimal routes are considered for execution. 
Once the mobile agent begins driving, the task cannot be altered, guaranteeing consistency and reliability in task completion.

\begin{algorithm}[t]
\caption{Initialization}\label{alg:planner}
\begin{algorithmic}[1]
\State $ N, \theta, p_t, s_0 $ 
\State $\mathcal{P} \gets \{p_t, s_0\}$
\While {$n < N$}
  \State ${\mathbf{W}} \gets \Phi(\mathcal{P})$ \Comment{LLM inference}
  \State $r, \tau \gets \mathcal{E}({\mathbf{W}})$ 
  \If{$ r, \tau > \theta $}
    \State \Return ${\mathbf{W}}$
  \EndIf
\EndWhile
\State \textbf{end}
\end{algorithmic}
\end{algorithm}

Algorithm~\ref{alg:planner} begins by initializing key parameters: the maximum number of iterations \( N \), the evaluation threshold \( \theta \), the target position \( p_t \), and the starting position \( s_0 \). A prompt \( \mathcal{P} \) is created, containing the task description and current position, which is then used by the LLM to generate waypoints. 
The LLM inference function \( \Phi \) produces a list of waypoints \( W \) based on this prompt, taking into account the grid map, current location, and required response format.
As the algorithm iterates, it evaluates the generated waypoints using the evaluation function \( \mathcal{E} \), which calculates the shortest path \( r \) and the number of turns \( \tau \). 
If the calculated path metrics \( r \) and \( \tau \) exceed the predefined threshold \( \theta \), the waypoint list is considered feasible and returned. 
This loop continues until a valid waypoint list is identified or the maximum number of iterations is reached. 
The algorithm ensures that only optimal routes are considered, thus providing a robust framework for waypoint generation and evaluation. 
This process incorporates global planning and rigorous waypoint evaluation to leverage LLM capabilities while ensuring safe and reliable path execution for mobile agents.

\subsection{Waypoint navigation}
After evaluating the waypoint list, the mobile agent begins to iterate through the waypoints. Due to potential sensor errors and the intricacies of the path-following method, it is essential for the mobile agent to appropriately select the following method. Simple waypoint following methods such as the dog curve and turn-and-drive can be employed to navigate the waypoints with a fixed distance. These methods enable the mobile agent to follow the sequence of waypoints with smooth and accurate navigation along the route.

In our approach, we decompose this procedure using a status transform matrix that maps the next driving command based on the current heading, current position, and the next waypoint. This matrix allows for dynamic adjustment and precise control during navigation. Additionally, the designed safety system ensures the execution is safe by preventing collisions with unknown obstacles. This is achieved using a position-sensitive detector and LIDAR beams, which continuously monitor the environment and provide real-time feedback for obstacle avoidance.

\begin{algorithm}[t]
\caption{Execution}\label{alg:follower}
\begin{algorithmic}[1]
\While {$w_i$ in $\mathbf{W}$}
  \State $s  \gets \mathcal{O}$ 
  \State $s' \gets \mathbf{W}$ 
  \State $ a \gets \Gamma(s, s') $ \Comment{Choose a following method}
  \State $ \Delta \gets ||s - s'|| $ 
  \If{$ \Delta < d $}
    \State \textbf{continue} 
  \EndIf
\EndWhile
\State \textbf{end}
\end{algorithmic}
\end{algorithm}

A algorithm~\ref{alg:follower} iterates over each waypoint \( w_i \) in the list \( W \). 
The current position \( s \) is updated using odometry data \( \mathcal{O} \), and the next waypoint \( s' \) is converted from the waypoint list \( W \). 
The following method is chosen based on the action command \( a \), which is determined by the selected path following function \( \Gamma(s, s') \). 
The distance \( \Delta \) between the current position \( s \) and the next waypoint \( s' \) is calculated.
If the distance \( \Delta \) is less than a predefined threshold \( d \), the algorithm continues to the next waypoint. 

\section{Experimental setup}

\subsection{Implement details}
This framework has been implemented on \textit{EyeBot simulator} \cite{braunl2023mobile}. 
The EyeBot simulator with virtual reality EyeSim VR is a multiple mobile robot simulator with VR functionality based on game engine Unity 3D that allows experiments with the same unchanged EyeBot programs that run on the real robots. 
We adjust the environmental values based on the task map from $5\times5$ to $11\times11$.
In each map, the mobile agent is at a random starting position and runs the proposed method in 10 episodes, and all performance metrics are averaged.
Three large language models are evaluated in the experiment including \textit{gpt-4o}, \textit{gemini-1.5-flash} and \textit{claude-3.5-sonnet} with the same system prompt and default temperature shown in Figure~\ref{fig:multilayer_framework}.

\subsection{Metrics}
We referenced the metrics from \cite{anderson_evaluation_2018} and \cite{zhao2021evaluation}, including success rate, average distance, and coverage rate.
The success rate indicates whether the paths generated by LLMs can cover the designated area. 
Average distance represents the average path length of the mobile robot, while coverage rate is a metric specific to coverage methods, used to assess the completeness of coverage path planning algorithms.

In traditional navigation evaluation standards, task termination is determined by the distance between the agent and the target point, which is effective for path planning problems with clearly defined start and end points. However, for coverage path planning algorithms, the generated paths do not have a clear endpoint, and the coverage path is autonomously decided by the LLM. Therefore, we have added a coverage rate metric to the comprehensive evaluation standards referenced from the cited sources. 
Inspired by Success weighted Path Length (SPL) from \cite{anderson_evaluation_2018}, we will refer to the following measure as CPL, short for \textbf{C}overage weighted by (normalized inverse) \textbf{P}ath \textbf{L}ength:

\begin{equation}
CPL = \frac{1}{N} \sum_{i=1}^{N} \frac{A_i}{\bar{A_i}} \frac{l_i}{max(p_i,l_i)}
\end{equation}

where $N$ means the number of test episodes. $A_i$ and $\bar{A_i}$ indicate the area of the coverage path and the area of the mission area, respectively.
The ratio of $A_i$ and $\bar{A_i}$ is expressed as the \textbf{C}overage \textbf{R}ate (CR), which is used to evaluate the completeness of the path.
The $l_i$ means the theoretical shortest path distance from the mobile agent start point, and the $p_i$ is the \textbf{P}ath \textbf{L}ength (PL) of the moving path by the agent.

\begin{table}[!t]
\centering
\setlength\tabcolsep{1.5pt}
\caption{Zero-shot coverage path planning performance using multiple LLM services in various environments } 
\begin{tabular}{|C{0.6in}|C{0.33in}C{0.27in}C{0.27in}|C{0.27in}C{0.27in}C{0.27in}|C{0.27in}C{0.27in}C{0.27in}|} 
    \hline 
    \multirow{2}{*}{Map Size} & \multicolumn{3}{c|}{GPT-4o} & \multicolumn{3}{c|}{Gemini-1.5} & \multicolumn{3}{c|}{Claude-3.5} \\
    \cline{2-10}
    {}           & CPL$\uparrow$& PL$\downarrow$& CR$\downarrow$& CPL   & PL   & CR     & CPL & PL & CR\\ \hline
    $5 \times 5$ & 0.95         & 34.2          & 96.4       & 0.87  & 44.5 & 87.8  & \textbf{0.99} & 37.2 &	100	  \\ \hline 
    $7 \times 7$ & 0.86         & 56.9          & 86.7       & 0.81	& 61.1 & 82.0   & \textbf{0.97} & 65.9 & 97.6  \\ \hline 
    $11\times11$ & 0.78         & 116           & 79.7       & 0.67	& 124  & 68.1	& \textbf{0.98} & 147  & 97.7  \\ \hline 
\end{tabular}
\label{tab:performance}
\end{table}

\subsection{Results and analysis}

The performance and time analysis are shown in Table~\ref{tab:performance} and Table~\ref{tab:delay_analysis}. All three models demonstrate the ability to plan a coverage path in a square space with a random start position. However, as the map size increases, the coverage rate decreases by approximately 5\% to 10\%, though all models maintain a coverage rate above 65\%. As shown in Table~\ref{tab:performance}, the model \textit{claude-3.5-sonnet} exhibits the best performance among the three models in terms of coverage rate and weighted path length. Changes in map size do not significantly affect the coverage rate and weighted path for the model \textit{gemini-1.5-flash}. Conversely, the model \textit{gpt-4o} achieves a higher coverage rate with smaller map sizes, but this rate decreases as the map size increases. As the map size grows, the actual path length increases more rapidly than the weighted path length, indicating that the planned paths include repeated visits to the same cells based on the random start position.

The differences in path length are attributed to the coverage rate of the planned path and the mobile agent's hardware capabilities, such as sensors and actuators. Since the evaluation processes locally with a short time cost (less than 300ms), we sum the inference time and the evaluation time as \( T_i \). \( T \) and \( T_d \) represent the total time spent and the driving part time cost, respectively.

\begin{table}[!t]
\centering
\setlength\tabcolsep{1.5pt}
\caption{Preceding and execution time analysis} 
\begin{tabular}{|C{0.6in}|C{0.27in}C{0.27in}C{0.27in}|C{0.27in}C{0.27in}C{0.27in}|C{0.27in}C{0.27in}C{0.27in}|} 
    \hline 
    \multirow{2}{*}{Map Size} & \multicolumn{3}{c|}{GPT-4o} & \multicolumn{3}{c|}{Gemini-1.5} & \multicolumn{3}{c|}{Claude-3.5} \\
    \cline{2-10}
    {}           & $T$   & $T_i$ & $T_d$ & $T$ & $T_i$& $T_d$ & $T$ & $T_i$ & $T_d$ \\ \hline
    $5 \times 5$ &  84.6 & 2.93  &  81.5 & 107 & 3.20 & 104 & 85.9	& \textbf{2.81} & 83.1   \\ \hline 
    $7 \times 7$ & 129   & 4.94  & 124   & 125 & 3.67 & 121 & 130	& \textbf{3.18} & 127  \\ \hline 
    $11\times11$ & 169   & 9.47  & 160   & 157 & \textbf{5.56} & 151 & 184	& {5.84} & 178  \\ \hline 
\end{tabular}
\label{tab:delay_analysis}
\end{table}

Model \textit{claude-3.5-sonnet} performs best and exhibits the fastest inference time in the experiment, planning fully coverage waypoints in various environments. Model \textit{gpt-4o} shows stable performance across different map sizes, demonstrating robustness and reliability. However, it is noted that the model's performance declines slightly as the map size increases, which could be attributed to the complexity of managing larger spaces and more waypoints. Model \textit{gemini-1.5-flash}, on the other hand, maintains consistent performance regardless of map size, although it occasionally introduces extra line break marks in its responses, which could be due to formatting issues within the LLM's output generation process.

Additionally, the path length differences highlight the varying capabilities of the mobile agents' hardware, such as sensor accuracy and actuator precision, which directly impact the execution of the planned paths. The evaluation process, which includes both inference and validation, ensures that the paths are not only feasible but also optimized for efficiency. 

Overall, the \textit{claude-3.5-sonnet} model excels in both performance and speed, making it ideal for scenarios requiring rapid and thorough coverage. The \textit{gpt-4o} model offers balanced performance with stability across various map sizes, making it a versatile choice. The \textit{gemini-1.5-flash} model, despite minor formatting issues, proves to be reliable with consistent performance. These insights can guide the selection of appropriate LLM services for specific coverage path planning tasks in mobile robotics.

\section{Discussion}
We propose a novel embodied framework for mobile agents, incorporating weighted evaluation metrics for the specific task of coverage path planning. A key factor of the framework is the use of zero-shot prompts to simplify LLM inference during the initial phase. This approach leverages the power of LLMs to generate effective waypoints without the need for extensive training data, thus streamlining the path-planning process.
During the navigation phase, we introduced a robust safety mechanism for mobile agents to avoid obstacles. This mechanism ensures that the mobile agents can navigate safely and efficiently in dynamic environments. Our experiments demonstrate that current LLMs have the capability to function as an embodied AI brain within mobile agents for specific tasks, such as area coverage, when guided by appropriately designed prompts.

The competition among LLM companies has significantly advanced the field, freeing researchers from the traditional labelling-training-validation loop in AI research. This shift allows for more focus on innovative applications and real-world deployment of AI technologies.
Future research will focus on evaluating path-planning problems in more realistic scenarios and simulation environments. This includes integrating more complex environmental variables and constraints to further evaluate and enhance the robustness of the proposed framework. Additionally, exploring the scalability of LLMs in diverse and larger-scale applications will be crucial in advancing the practical deployment of embodied AI systems in mobile robotics.

\section*{ACKNOWLEDGMENT}
The authors would like to thank all \xblackout{the Renewable Energy Vehicle Project (REV) sponsors} for their support on this project, especially \xblackout{Stockland, Allkem and CD Dodd}. 

\bibliographystyle{IEEEtran}
\bibliography{ref}

\end{document}